\documentclass[runningheads]{llncs}
\usepackage[T1]{fontenc}

\usepackage[table,xcdraw]{xcolor}
\usepackage{graphicx}
\usepackage{amsfonts} 
\usepackage{bm} 
\usepackage{colortbl} 
\usepackage[table]{xcolor} 
\usepackage{amsmath}
\usepackage{amssymb}
\usepackage{booktabs}
\usepackage{algorithm}
\usepackage{algorithmic}
\usepackage{enumitem}
\definecolor{grayish}{RGB}{240, 240, 240} 
\definecolor{grayish_2}{RGB}{230, 230, 230} 
\usepackage{hyperref}
\usepackage{adjustbox}
\hypersetup{
    colorlinks=true,
    linkcolor=blue,
    citecolor=blue,
    urlcolor=blue
}
\usepackage{multirow}
\usepackage[table]{xcolor}
\begin{document}
\title{FissionFusion: Fast Geometric Generation and Hierarchical Souping for Medical Image Analysis}
\titlerunning{FissionFusion}

\author{Santosh Sanjeev,
Nuren Zhaksylyk,
Ibrahim Almakky, Anees Ur Rehman Hashmi, Mohammad Areeb Qazi \and Mohammad Yaqub}
\authorrunning{S. Sanjeev et al.}
%
\institute{Mohamed bin Zayed University of Artificial Intelligence, Abu Dhabi, UAE \\
\email{\{firstname.lastname\}@mbzuai.ac.ae} \\
}
%
%

%
%
\maketitle              
\begin{abstract}

The scarcity of well-annotated medical datasets requires leveraging transfer learning from broader datasets like ImageNet or pre-trained models like CLIP. Model soups averages multiple fine-tuned models aiming to improve performance on In-Domain (ID) tasks and enhance robustness against Out-of-Distribution (OOD) datasets. However, applying these methods to the medical imaging domain faces challenges and results in suboptimal performance. This is primarily due to differences in error surface characteristics that stem from data complexities such as heterogeneity, domain shift, class imbalance, and distributional shifts between training and testing phases. To address this issue, we propose a hierarchical merging approach that involves local and global aggregation of models at various levels based on models' hyperparameter configurations. Furthermore, to alleviate the need for training a large number of models in the hyperparameter search, we introduce a computationally efficient method using a cyclical learning rate scheduler to produce multiple models for aggregation in the weight space. Our method demonstrates significant improvements over the model souping approach across multiple datasets (around 6\% gain in HAM10000 and CheXpert datasets) while maintaining low computational costs for model generation and selection. Moreover, we achieve better results on OOD datasets than model soups. The code is available at \url{https://github.com/BioMedIA-MBZUAI/FissionFusion}.


\keywords{Model Soups \and Medical Image Analysis \and Model Merging \and Transfer Learning}
\end{abstract}
\section{Introduction}
Deep learning (DL) has emerged as the de facto standard for various computer vision tasks, consistently achieving state-of-the-art performance. A pivotal contributor to this success lies in the availability of pre-trained models.  In recent years, a well-established paradigm has emerged: pre-training models on large-scale data, such as ImageNet \cite{imagenet} followed by fine-tuning on target tasks with limited training data \cite{transfer_learning_1,transfer_learning_2}. This strategy has proven particularly effective in domains with constrained data availability, such as medical imaging. The scarcity of annotated medical datasets, coupled with the challenges of data acquisition and ethical considerations, highlights the importance of transfer learning from large-scale datasets. Although public data in medical imaging is increasing, it is usually smaller in size compared to natural image databasets, which has led to the widespread adoption of transfer learning from ImageNet \cite{imagenet}, to improve performance on medical tasks \cite{medical_imagenet_1,medical_imagenet_2,medical_imagenet_3}.
\begin{figure}[t]
\centering
\includegraphics[width=1\textwidth]{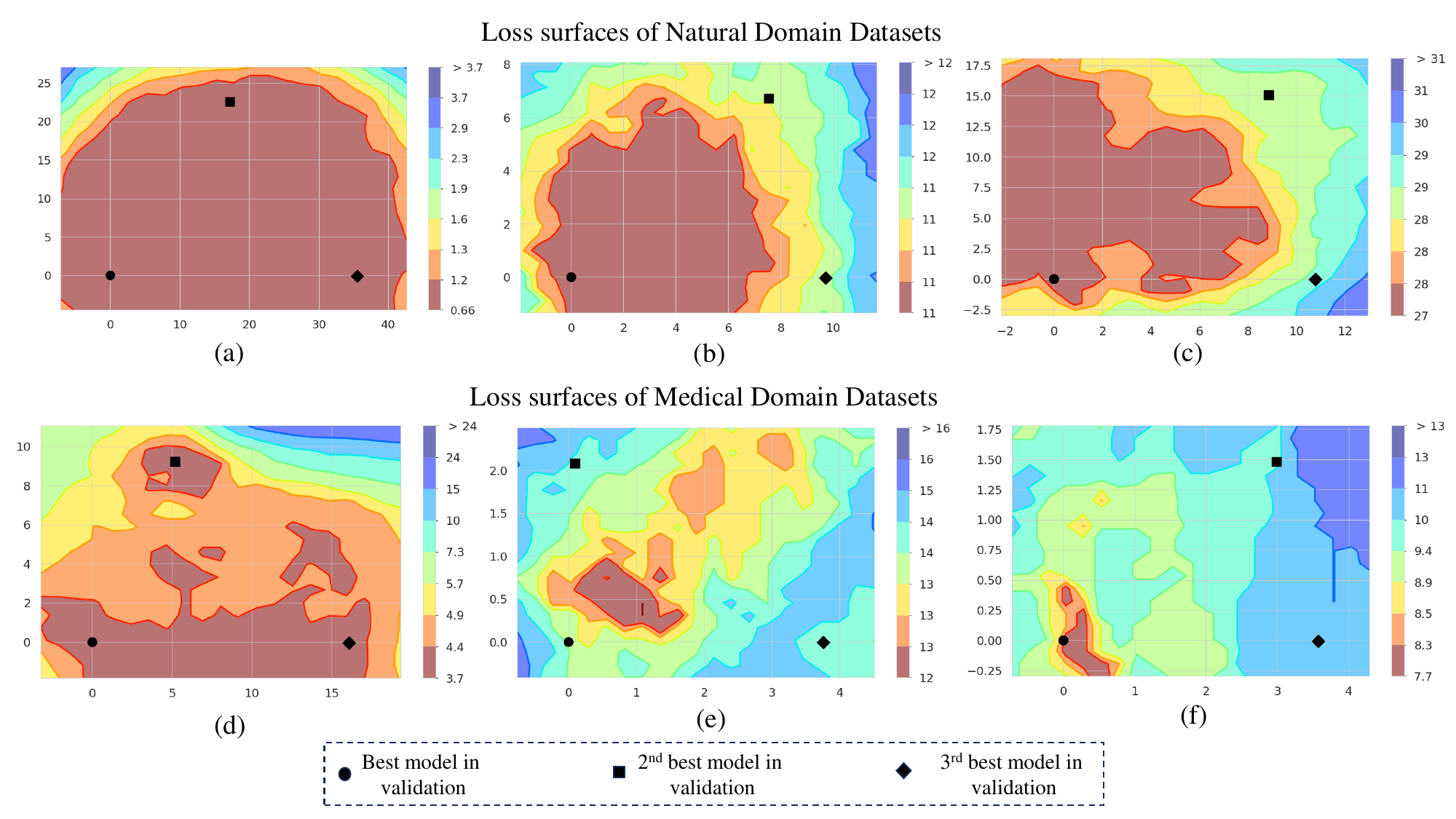}
\caption{This figure illustrates the validation error on a two-dimensional slice of the error landscapes for various natural and medical domain datasets following an approach outlined in \cite{garipov2018loss}. (a) CIFAR-10 \cite{krizhevsky2009learning} (b) CIFAR-100 \cite{krizhevsky2009learning} (c) FGVC-Aircrafts \cite{maji13fine-grained} (d) RSNA Pneumonia  \cite{rsna} (e) APTOS \cite{aptos2019-blindness-detection} (f) HAM10000 \cite{DVN/DBW86T_2018}. We employ the 3 best-performing models from the validation set, with the best model serving as the reference (origin).} 
\label{fig:loss_curves}
\end{figure}

The common practice in transfer learning is to adapt a pre-trained model to a downstream task by fine-tuning. This involves conducting a grid search to explore various hyperparameter combinations and selecting the model that performs best on the validation set. Another approach is employing ensemble techniques \cite{ensemble}, where multiple models are utilized simultaneously, albeit at the expense of increased computational and memory requirements, especially at inference time. Recent research by \cite{neyshabur2020being} has noted that fine-tuned models optimized independently from the same pre-trained initialization often converge to similar error basins. A previous work \cite{swa} has demonstrated that averaging weights along a single training trajectory can enhance model performance in non-transfer settings. Motivated by these observations, \cite{wortsman_model_2022} proposed model averaging - Model Soups (MS) as an alternative approach to ensembles, aiming to derive a single model that achieves a good performance in terms of accuracy and inference time on In-Domain (ID) as well as Out-of-Distribution (OOD) datasets. They show that averaging several fine-tuned models trained with different hyperparameter settings results in a better model. This method is particularly effective in scenarios with natural imaging datasets or those without significant domain shifts. Several other works have improved upon the model merging concept by adopting Fisher-based weight averaging \cite{weightaveraging}, Task Arithmetic \cite{taskarithmetic}, Pruning \cite{zimmer2023sparse}, Gradient-based Matching \cite{modelmerginggradientbased}, Tie-Merging \cite{tiesmerging}, and FedSoups \cite{fedsoups} for Federated Learning setting. 

However, complexities arise when dealing with medical datasets. As illustrated in Fig.  \ref{fig:loss_curves}, a significant contrast arises between the error surfaces of natural imaging datasets (a-c), characterized by smoothness and convexity, and those of medical datasets (d-f), exhibiting pronounced roughness and the presence of multiple local optima within the global minima. This roughness is attributed to various intricacies inherent in medical data, including data heterogeneity, domain shift from pre-trained (ImageNet) networks, class imbalance, and distribution shift between the training and testing phases. Such complexities often result in challenging optimization landscapes, where models are prone to getting stuck in suboptimal minima. Consequently, the effectiveness of model averaging is compromised, often leading to subpar performance outcomes. Few works have adopted model souping for medical datasets \cite{modelsoupsmedical_1,modelsoupsmedical_4,modelsoupsmedical_5,modelsoupsmedical_6}. Most of these studies have merely applied uniform or greedy souping to few models, and the majority have not explored or analyzed model souping from the perspective of error surfaces. Additionally, the process of model averaging typically involves training multiple models with different hyperparameter settings, which can be computationally intensive. In contrast to previous research, our work explores computationally efficient model generation and averaging in a transfer learning context, particularly to domains with significant domain shifts.


In Model Soups \cite{wortsman_model_2022}, the process of fine-tuning consists of two main steps: (1) Model generation - fine-tuning models with various hyperparameter configurations, and (2) Model selection - selecting the model with the highest accuracy on the held-out validation set and then performing either uniform or greedy souping. In our study, we explore the challenges associated with both steps in the medical image analysis domain. Drawing from insights in \cite{neyshabur2020being} and \cite{garipov2018loss}, we propose a Fast Geometric Generation (FGG) approach to efficiently generate models with minimal computational overhead. Additionally, we address the model selection process by introducing Hierarchical Souping (HS), which is better suited for medical data because of the aforementioned complexities. Our key contributions are as follows:
\begin{itemize}
\item We propose \textbf{Fast Geometric Generation (FGG)} approach, which uses cyclical learning rate scheduler in the weight space for efficient model generation. This approach achieves superior results compared to model soups at a lower computational cost.
\item We introduce a novel selection mechanism - \textbf{Hierarchical Souping} (HS), tailored specifically for the medical image analysis domain, which performs model averaging at different levels. 
\item We perform a comprehensive analysis of model souping across various datasets in both natural and medical domains. We demonstrate that combination of FGG and HS for selection significantly enhance results, improve robustness and increase generalization to out-of-distribution datasets.

\end{itemize}




\begin{figure}[t]
    \centering
    \includegraphics[width=1.1\linewidth]{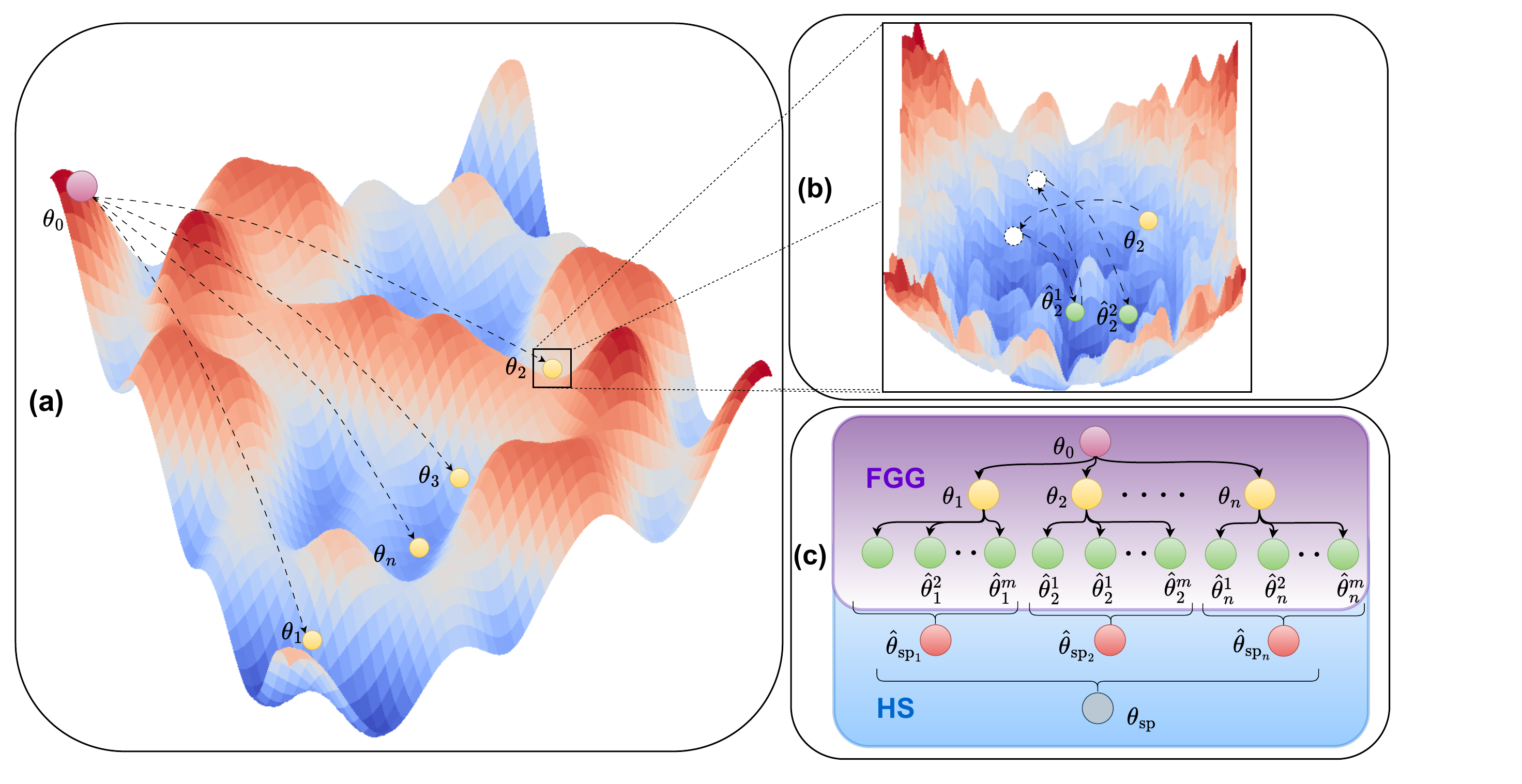}
    \caption{An illustration of (a) Loss landscape of fine-tuned models (b) Fast Geometric Generation(FGG) approach using cyclical learning rate scheduler (c) FGG and the Hierarchical Souping (HS) approach.}
    \label{fig:methodology}
\end{figure}

\section{Methodology}
Let \( \mathcal{D}_\text{train} = \{(x_i, y_i)\}_{i=1}^N \) denote the training dataset where \( x_i \in \mathbb{R}^{d} \) represents the input data and \( y_i \in \{1, 2, \ldots, C\} \) denotes the corresponding label from a set of \textit{C} classes. Similarly, let \( \mathcal{D}_\text{val} = \{(x_j, y_j)\}_{j=1}^M \) be the validation dataset. We adapt a pre-trained model to our task by performing linear probing for a few epochs as a warm-up step, where we only update the weights of the last layer while freezing the rest of the model. Let \( \theta_\text{0} \) denote the parameters of the linear probed model. As mentioned above, the souping procedure involves two main steps (1) Generation of the models by fine-tuning (2) Selection of models that perform best on the \( \mathcal{D}_\text{val} \). The illustration of our proposed approaches (FGG and HS) can be seen in Fig. \ref{fig:methodology}.
\newline

\noindent \textbf{Fast Geometric Generation (FGG).} We design FGG to generate a range of diverse neural networks while navigating through the weight space taking small steps. The primary objective is to explore variations in network weights without straying into regions likely to lead to low test performance. Instead of employing the conventional hyperparameter grid search, we opt for a more focused strategy. In FGG, we only iterate through a single hyperparameter i.e. (learning rate), and the rest of the hyperparameters are kept constant. This helps reduce the hyperparameter search space and the computational cost. 

In FGG, we start with \( \theta_\text{0} \) as an initialization and fully train the model using $n$ different learning rates, resulting in a set of parameters $\Theta = \{\theta_1, \dots, \theta_n\}$. After obtaining $\Theta$, we implement the \textbf{Fission} process. We initialize the weights $ \theta_t \in \Theta$, and carry out a second training process for a set of iterations $\textit{I} = \{1, \dots, k\}$ to generate a set of parameters $\hat{\Theta} = \{\hat{\theta}^1, \dots, \hat{\theta}^m\}$. During this, we employ a learning rate scheduler to cyclically change the learning rate every cycle $c$, where $c$ is defined as a set of number of iterations and is an even number. This encourages \( \theta_t \) to diverge from its initialization, while maintaining validation accuracy. We use a function $\alpha(i)$ to control the learning rate at iteration \( i \in \textit{I} \) as follows:

\[
\alpha(i) = \begin{cases}
    \alpha_2 . (2t(i)) + \alpha_1(1 - 2t(i)) & \text{if } 0 < t(i) \leq 0.5, \\
    \alpha_1 . (2t(i) - 1) + \alpha_2(2 - 2t(i)) & \text{if } 0.5 < t(i) \leq 1,
\end{cases}
\]

where \( t(i) \)\( = \)\( \frac{1}{c}(\text{mod}(i - 1, c) + 1) \), $\alpha_2$ and $\alpha_1$ (\(\alpha_2\) \(\leq \) \(\alpha_1\)) are hyperparameters used to control the minimum and maximum learning rates, respectively. Using $\alpha(i)$, we train the network \( \theta_t \) to form the new set of parameters $\hat{\Theta}$, where each $\hat{\theta} \in \hat{\Theta}$ is the model parameters when the learning rate reaches its minimum value, \( \alpha(i) = \alpha_2 \). This occurs at the point where \( \text{mod}(i - 1, c) + 1 = \frac{c}{2}\) and \( t(i) = \frac{1}{2} \). 
We follow this training process for every $\theta_t \in \Theta$, which results in a $\{\hat{\Theta}_1,\dots,\hat{\Theta}_n\}$ .


During high learning rate intervals (close to \( \alpha_1 \)), the weight \( \theta_t \) traverses the weight space with larger strides, potentially leading to higher test error. Conversely, in low learning rate episodes, \( \theta_t \) transitions to smaller steps, reducing test error. This mechanism facilitates incremental movements in weight space, aiding models in evading local minima. Additionally, it gathers diverse models for averaging, reducing the necessity for an extensive grid search.
\newline



\noindent \textbf{Hierarchical Souping (HS).} In \cite{wortsman_model_2022}, the greedy souping approach outperforms uniform averaging by sequentially adding models to the soup if they improve accuracy on \( \mathcal{D}_\text{val} \). This approach can yield suboptimal results when the best model is stuck in a local optimum because of the uneven error surface caused due to domain shift. To solve this, we propose Hierarchical Souping, where models are merged at different levels. Starting from the parameter sets $\Theta$ and $\hat{\Theta}$ acquired during the FGG phase, we adopt a local souping approach where the generated models $\{\hat{\Theta}_1,\dots,\hat{\Theta}_n\}$ are averaged along with the corresponding initialization $\theta_t$ (greedy or uniform i.e. \( {\hat{\theta}}_{\text{sp}_t} = \frac{1}{m+1} [\sum_{i=1}^{m} \hat{\theta}_t^i +\theta_t] \) ) at different levels resulting in  
$ \{ \hat{\theta}_{{sp}_1}, 
     \hat{\theta}_{{sp}_2}, \dots,
     \hat{\theta}_{{sp}_n} \} $ 
and a greedy averaging technique is used at the top level giving \({\theta}_{\text{sp}} \) (GoG refers to Greedy at all levels whereas GoU refers to Greedy at the top level and Uniform at the lower levels). This approach enables networks to escape local minima at the lower levels by local aggregation, giving a good subset of generalizable models that can be averaged in a greedy manner at the top level.

\section{Experiments}
\textbf{Implementation Details.} Our research investigates two model architectures: DeiT-B (Transformer) and ResNet50 (CNN), both pre-trained on ImageNet. Before full fine-tuning, we conduct a 10-epoch warmup where the whole network except the last layer is frozen (\( \theta_\text{0} \)). We then unfreeze the entire model and perform a full fine-tuning known as the LP-FT approach \cite{kumar2022finetuning}. We use AdamW optimizer with a cosine annealing scheduler for 50 epochs, batch size of 128, and an image resolution of 224$\times$224. 

For the model soups experiments, we perform a grid search over learning rate, seeds, and augmentations similar to \cite{wortsman_model_2022} for the general grid search (GS) experiments. Learning rates (LR) include (1e-3, 5e-4, 1e-4, 5e-5, 1e-5, 5e-6, 1e-6, 1e-7) and Augmentations include minimal (random crop covering 90-100\% of image), medium (default timm library settings \cite{rw2019timm}), and heavy (RandAugment with $N=2$, $M=15$) \cite{cubuk2019randaugment}. Each hyperparameter configuration is run with two seeds, resulting in 48 models in total for baseline results \cite{wortsman_model_2022}. 

 For our method, we focuses solely on the learning rate as the varying hyperparameter, where we fine-tune eight models initially. Each model is trained with a different learning rate, employing heavy augmentation and a fixed seed. According to \cite{wortsman_model_2022,moussa2022hyperparameter}, the learning rate emerges as the most influential parameter, positioning the model in the loss surface. All experiments use the AdamW optimizer with a cyclical learning rate scheduler, where the learning rate ranges between \( \alpha_1=1e-5 \) and \( \alpha_2=1e-8 \). From each base model, we generate five models using FGG, resulting in a total of 48 models. We train the eight initial models for 50 epochs and then we perform the second stage training process (FGG) for 17 (4 epochs per cycle) epochs with cyclical learning rate collecting 5 models per base model, at the end of the training process.  We also provide an analysis and justification for choosing learning rate as a top-level hyperparameter using Linear Mode Connectivity (LMC) in Fig. 1 in \ref{sec:app_sec_b} in the Appendix.
 \newline
\textbf{Datasets.} For the primary experiments, we consider two natural imaging domain datasets and five medical domain datasets. We use CIFAR10 \cite{krizhevsky2009learning} and CIFAR100 \cite{krizhevsky2009learning}, partitioning the training dataset into train/validation sets in a 90\%:10\% ratio. We utilize the official test split provided. For the CheXpert \cite{irvin2019chexpert}  and HAM10000 \cite{DVN/DBW86T_2018} datasets, we adhere to the official train/validation/test splits. For APTOS \cite{aptos2019-blindness-detection}, EyePACs \cite{eyepacs}, and RSNA-Pneumonia \cite{rsna} datasets, we split the data into train/validation/test sets in an 80\%:10\%:10\% ratio, given that only the training dataset was publicly available. It is noteworthy that all datasets are multiclass, except CheXpert, which is a multilabel dataset. For the OOD experiments, we consider the CIFAR10.1 \cite{recht2018cifar10.1,torralba2008tinyimages} having 2000 test samples from the natural imaging domain. For the medical imaging domain, we use the Messidor and Messidor-2 \cite{ImageAnalStereol1155} datasets, sampling 10\% of the data for the test set. We also use the MIMIC-CXR \cite{Johnson2019Dec} dataset and follow its official test split. 



\section{Results and Discussion}
\begin{table}[t]
\centering

\caption{Performance comparison of different methods. (GS (best) - best model on the validation set from Grid Search,  FGG(best) - best model on the validation set from Fast Geometric Generation, GoU - Greedy of Uniform, GoG - Greedy of Greedy), Bold numbers mean best and underlined are the second best}
\label{table:main_results}

{\renewcommand{\arraystretch}{1.1}
\resizebox{\textwidth}{!}{
\begin{tabular}{c|c|c|c|c|c|c|c|c}
\hline
\textbf{\begin{tabular}[c]{@{}c@{}}Model \end{tabular}} & \textbf{Method}     & \textbf{\begin{tabular}[c]{@{}c@{}}CIFAR10\\ (Acc.)\end{tabular}} & \textbf{\begin{tabular}[c]{@{}c@{}}CIFAR100\\ (Acc.)\end{tabular}} & \textbf{\begin{tabular}[c]{@{}c@{}}APTOS\\ (F1)\end{tabular}} & \textbf{\begin{tabular}[c]{@{}c@{}}HAM10000\\ (Recall)\end{tabular}} & \textbf{\begin{tabular}[c]{@{}c@{}}RSNA\\ (F1)\end{tabular}} & \textbf{\begin{tabular}[c]{@{}c@{}}CheXpert\\ (AUC)\end{tabular}} & \textbf{\begin{tabular}[c]{@{}c@{}}EyePACs\\ (F1)\end{tabular}} \\ \hline
\multirow{6}{*}{\rotatebox{90}{ResNet50}}                                      & GS(best)     & 0.9769                                                                 & 0.8380                                                                  & 0.7086                                                         & 0.6074                                                                & 0.9444                                                        & 0.8444       & 0.4750                                                      \\
                                                               & Uniform Soup        & 0.8703                                                                 & 0.7652                                                                  & 0.5509                                                         & 0.5698                                                                & 0.9171                                                        & 0.5752 & 0.1738                                                             \\
                                                               & Greedy Soup         & 0.9769 & 0.8401 & \textbf{0.7247} & 0.6074 & 0.9444 & \underline{0.8444} & 0.4750 \\
                                                               & FGG(best)    & 0.9783                                                                 & \underline {0.8464}                                                            & 0.7172                                                         & 0.6614                                                                & 0.9518                                                        & 0.8434 & 0.4874                                                             \\ \cline{2-9} 
                                                               & \textbf{GoU (Ours)} & \textbf{0.9785}                                                        & 0.8457                                                                  & 0.6909                                                         & \textbf{0.6818}                                                       & \textbf{0.9545}                                               & 0.8351  & \textbf{0.4905}                                                          \\
                                                               & \textbf{GoG (Ours)} & \textbf{0.9785}                                                        & \textbf{0.8477}                                                         & \underline {0.7172}                                                   & \underline {0.6614}                                                          & \underline {0.9518}                                                  & \textbf{0.8488}   & \underline{0.4900}                                                 \\ \hline
\multirow{6}{*}{\rotatebox{90}{DeiT-B}}                                         & GS(best)     & 0.9871                                                                 & 0.8919                                                                  & 0.6903                                                         & 0.6487                                                                & 0.9503                                                        & 0.8143          & 0.4807                                                   \\
                                                               & Uniform Soup        & 0.9386                                                                 & 0.8551                                                                  & 0.1637                                                         & 0.1429                                                                & 0.4147                                                        & 0.7177  & 0.1697                                                            \\
                                                               & Greedy Soup         & 0.9892                                                                 & 0.8968                                                                  & 0.6785                                                         & \underline{0.6487}                                                                & 0.9503                                                        & 0.8068  & 0.4865                                                            \\
                                                               & FGG(best)     & 0.9876                                                                 & 0.8963                                                                  & \textbf{0.7011}                                                & 0.6393                                                                & 0.9529                                                        & \underline {0.8619}  & \textbf{0.5029}                                                     \\ \cline{2-9} 
                                                               &  \textbf{GoU (Ours)} & \underline {0.9899}                                                           & \underline {0.8968}                                                            & 0.6976                                                         & \textbf{0.6495}                                                       & \textbf{0.9579}                                               & 0.7609   & 0.4903                                                          \\
                                                               & \textbf{GoG (Ours)} & \textbf{0.9901}                                                        & \textbf{0.8987}                                                         & \underline {0.7003}                                                   & 0.6393                                                          & \underline {0.9529}                                                  & \textbf{0.8644} & \underline{0.4940}                                                    \\ \hline
\end{tabular}}}
\end{table}




\begin{figure}[t]
    \centering
    \includegraphics[width=\linewidth]{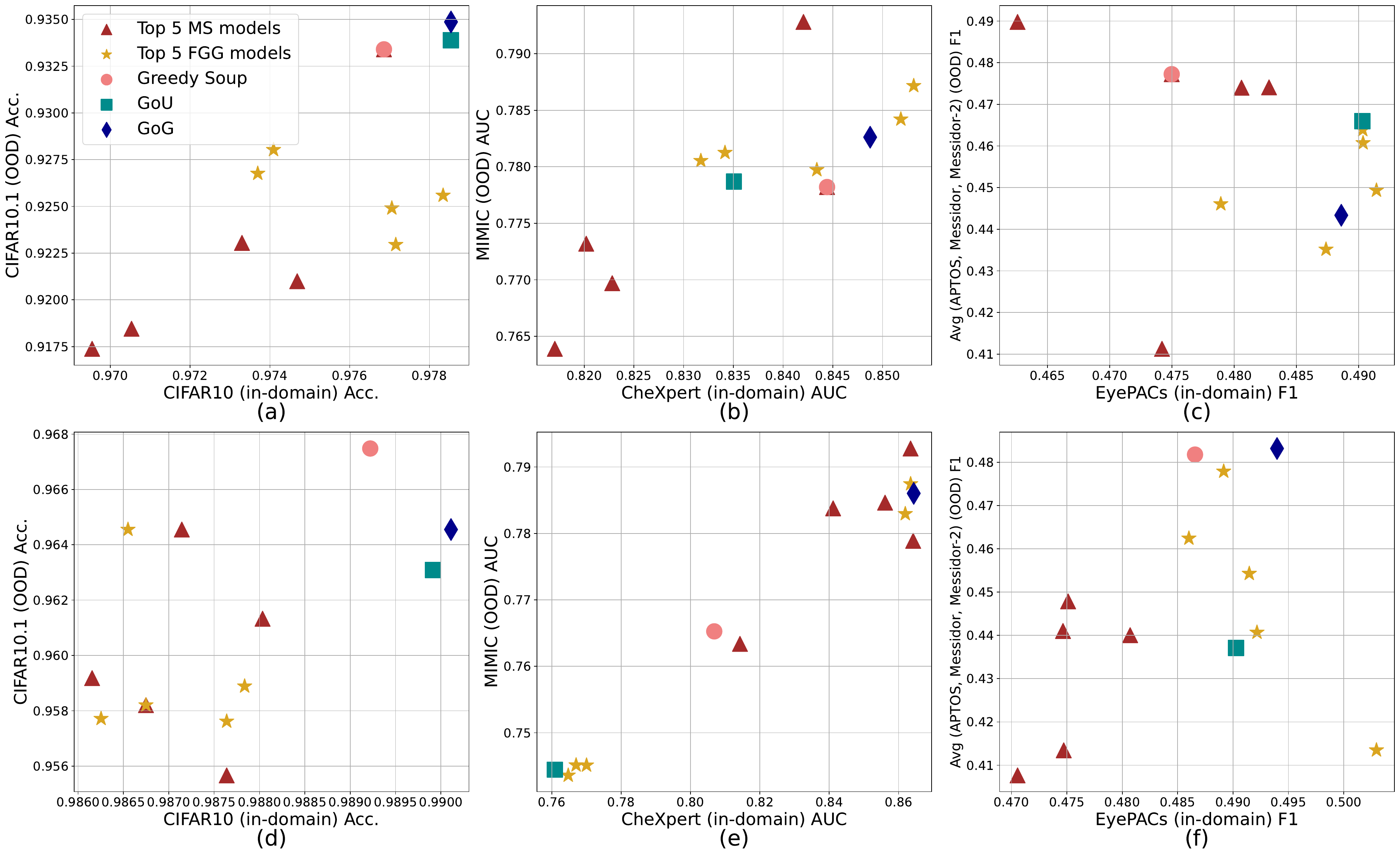}
    \caption{OOD analysis for different architectures on various datasets (a) CIFAR10 v/s CIFAR10.1 - ResNet50 (b) CheXpert v/s MIMIC - ResNet50 (c) APTOS v/s (EyePacs, Messidor, Messidorv2) - ResNet50 (d)  CIFAR10 v/s CIFAR10.1 - DeiT-B (e) CheXpert v/s MIMIC - DeiT-B (f) APTOS v/s (EyePacs, Messidor, Messidorv2) - DeiT-B. We do not plot the results of Uniform Soups as it performs poorly.}
    \label{fig:ood}
\end{figure}

As seen in Table \ref{table:main_results}, we observe that our method achieves better results both in the natural as well as medical imaging domains. In CIFAR datasets, we observe that GoU and GoG achieve better results compared to model soups, though it is not significant due to the smooth convex error surface. However, when testing on medical datasets, we observe around \textbf{6\%} improvement in Recall (GoG) for the HAM10000 dataset (ResNet50) as well as around \textbf{6\%} gain (GoG) in AUC(DeiT-B) for the CheXpert dataset. Both the GoU as well as GoG approaches achieve similar results except in the case of CheXpert (DeiT-B), where GoU significantly lags behind GoG. The GoG is always better than the Greedy Soup approach in almost all cases and is achieved at a significantly lower computational cost, as we do not have to perform the full grid search. Greedy soups does not perform as expected when fine-tuning on medical datasets as the averaging starts from the best model, which can be an issue due to uneven error surfaces. For example, for the DeiT-B on the CheXpert dataset, we observe that the best model has a high validation score, but a low test score indicating its not generalizable as it might have been stuck in a local minima. The FGG process overcomes this issue through its ability to escape numerous local minima, while the acquired models cluster around the same area in the error surface, facilitating smoother averaging. The local averaging in HS approach allows smoother averaging at local surfaces and greater diversity at the top level.  We further conduct an ablation study in Table \ref{tab:ablation} in Section \ref{sec:app_sec_a} in the Appendix, exploring greedy souping on the FGG models, and HS on the grid-search generated models. \\

\noindent \textbf{OOD Analysis.} We also conduct an analysis on OOD scenarios for both natural and medical domain datasets. Performance comparison of souped models on OOD data is important as averaged models should demonstrate robustness to distribution shifts \cite{wortsman_model_2022}. In Fig. \ref{fig:ood}, we compare the performance of various approaches in both ID and OOD datasets, where the top five models are selected based on validation set results. In almost all the cases, either GoU or GoG yields similar or better results compared to greedy soups in both ID and OOD tasks at a much lower computational cost. Our approach results in higher ID and OOD performance gains, particularly for medical imaging datasets, which can be attributed to the FGG approach aiding models in escaping local minima. HS also contributes to this, by facilitating easier averaging between models in the error surface.

\section{Conclusion}
This work investigates challenges associated with model averaging in transfer learning settings, particularly in medical imaging domain where significant domain shifts can occur. To address this, we introduce Fast Geometric Generation (FGG) leveraging hyperparameter significance and a cyclical learning rate scheduler. Moreover, we propose a Hierarchical Souping mechanism, which involves averaging models at different levels based on the smoothness of the error surface and hyperparameter significance. The proposed generation and selection methodologies yield notable performance enhancements compared to the traditional model souping approach. While our work achieves improved results, we observe instances where models from grid search occasionally outperform averaged models due to very rough error landscapes. This suggests a potential improvement for enhancing generalizability by smoothing the error surface, an aspect we plan to focus on in future endeavors.

\bibliographystyle{splncs04}
\bibliography{references}

\pagebreak
\appendix

\newpage

\appendix
\section*{Appendix}
\vspace{-5pt}

\renewcommand{\thesection}{A\arabic{section}}
\setcounter{section}{0}
\setcounter{figure}{0}
\setcounter{table}{0}

\section{Ablation study for different generation and selection techniques}
\vspace{-15pt}
\label{sec:app_sec_a}
\begin{table}[h]
\vspace{-8pt}
\centering
\caption{Ablation study exploring the Performance comparison of different combinations. (GS - Grid search for model souping, GoU - Greedy of Uniform , GoG - Greedy of Greedy),Bold numbers mean best and underlined are the second best}
\label{tab:ablation}

\resizebox{\textwidth}{!}{
\begin{tabular}{c|c|cccccc}
\hline
\textbf{\begin{tabular}[c]{@{}c@{}}Model\end{tabular}} & \textbf{Method}  & \textbf{\begin{tabular}[c]{@{}c@{}}CIFAR10\\ (Acc.)↑\end{tabular}} & \textbf{\begin{tabular}[c]{@{}c@{}}CIFAR100\\ (Acc.)↑\end{tabular}} & \textbf{\begin{tabular}[c]{@{}c@{}}APTOS\\ (F1)↑\end{tabular}} & \textbf{\begin{tabular}[c]{@{}c@{}}HAM10000\\ (Recall)↑\end{tabular}} & \textbf{\begin{tabular}[c]{@{}c@{}}RSNA\\ (F1)↑\end{tabular}} & \textbf{\begin{tabular}[c]{@{}c@{}}CheXpert\\ (AUC)↑\end{tabular}} \\ \hline
\multirow{5}{*}{ResNet50}                                      & GS+HS(GoU)       & 0.9777                                                                 & 0.8434                                                                  & 0.6553                                                         & 0.6589                                                                & 0.9474                                                        & 0.8194                                                             \\
                                                               & GS+HS(GoG)       & 0.9771                                                                 & 0.8409                                                                  & 0.7045                                                         & 0.6071                                                                & 0.9444                                                        & 0.8444                                                             \\
                                                               & FGG+ UniformSoup & 0.9218                                                                 & 0.8076                                                                  & 0.5912                                                         & 0.6070                                                                & 0.9104                                                        & 0.7313                                                             \\
                                                               & FGG + GreedySoup & 0.9776                                                                 & 0.8466                                                                  & 0.7172                                                         & 0.6614                                                                & 0.9518                                                        & 0.8488                                                             \\
                                                               & FGG+HS(Aug@top)  & 0.9769                                                                 & 0.8410                                                                  & 0.7118                                                         & 0.6317                                                                & 0.9444                                                        & 0.8444                                                             \\ \hline
\multirow{5}{*}{DeiT-B}                                        & GS+HS(GoU)       & 0.9887                                                                 & 0.8975                                                                  & 0.6797                                                         & 0.6713                                                                & 0.9499                                                        & 0.8609                                                             \\
                                                               & GS+HS(GoG)       & 0.9886                                                                 & 0.8957                                                                  & 0.6990                                                         & 0.6487                                                                & 0.9530                                                        & 0.8707                                                             \\
                                                               & FGG+ UniformSoup & 0.9309                                                                 & 0.8509                                                                  & 0.3067                                                         & 0.1429                                                                & 0.3781                                                        & 0.7199                                                             \\
                                                               & FGG + GreedySoup & 0.9895                                                                 & 0.9003                                                                  & 0.6828                                                         & 0.6833                                                                & 0.9529                                                        & 0.8653                                                             \\
                                                               & FGG+HS(Aug@top)  & 0.9892                                                                 & 0.8946                                                                  & 0.6741                                                         & 0.6487                                                                & 0.9503                                                        & 0.8691                                                             \\ \hline
\end{tabular}}
\end{table}
\vspace{-25pt}

\renewcommand{\thesection}{B\arabic{section}}
\setcounter{section}{0}
\setcounter{figure}{0}

\section{Linear Mode Connectivity} \label{sec:app_sec_b}
\vspace{-30pt}

\begin{figure}[h]
    \centering
    \includegraphics[width=0.9\linewidth]{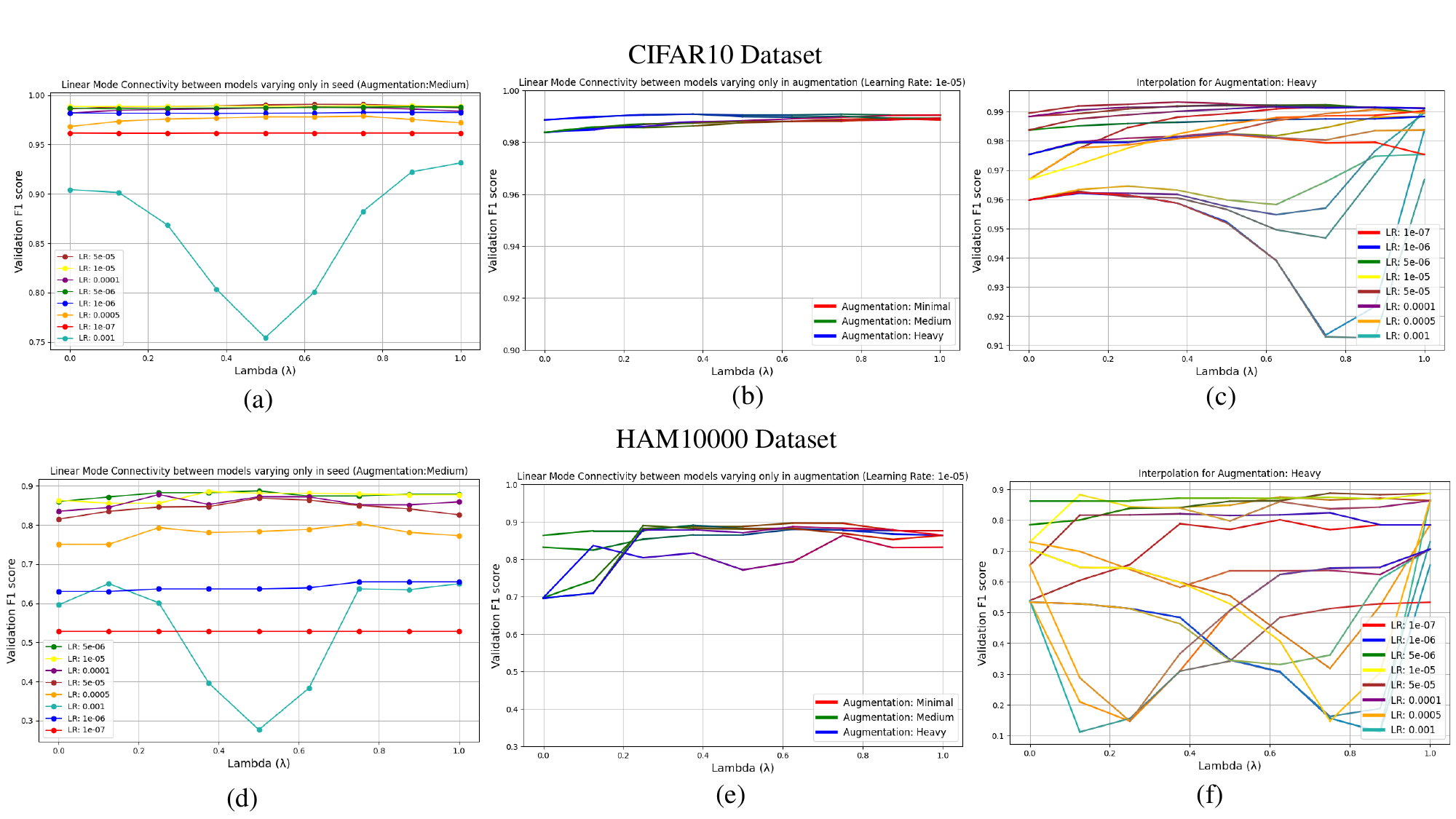}
    \vspace{-12pt}    \caption{Hyperparameter Analysis using Linear Mode Connectivity (LMC) (\(\theta = \lambda \cdot \theta_{A} + (1 - \lambda) \cdot \theta_{B}\)), where \(\theta_{A}\) and \(\theta_{B}\) differ only in one hyperparameter. (a) and (d) LMC between models varying only in seed. (b) and (e) LMC between models varying only in augmentation. (c) and (f) LMC between models varying only in learning rate.}
    \label{fig:lmc}
\end{figure}
\vspace{-30pt}

\end{document}